\documentclass[conference]{IEEEtran}
\usepackage{times}

\usepackage[numbers,sort&compress]{natbib}
\usepackage{multicol}
\usepackage[bookmarks=true]{hyperref}
\usepackage{etoolbox}
\usepackage[font=large,skip=0pt]{caption}

\usepackage{graphicx}

\begin{document}

% paper title
\title{Usability Squared: Principles for doing good systems research in robotics}
% You will get a Paper-ID when submitting a pdf file to the conference system
%\author{Author Names Omitted for Anonymous Review.}

\author{\authorblockN{Soham Sankaran}
\authorblockA{Cornell University\\
Email: soham@soh.am}
\and
\authorblockN{Ross A. Knepper}
\authorblockA{Cornell University\\
Email: rak@cs.cornell.edu}
}
%\patchcmd{\@maketitle}{\addvspace{0.5\baselineskip}\egroup}{\addvspace{-1\baselineskip}\egroup}{}{}

\maketitle
\begin{abstract}
Despite recent major advances in robotics research, massive injections of capital into robotics startups, and significant market appetite for robotic solutions, large-scale real-world deployments of robotic systems remain relatively scarce outside of heavy industry and (recently) warehouse logistics. In this paper, we posit that this scarcity comes from the difficulty of building even merely functional, first-pass robotic applications without a dizzying breadth and depth of expertise, in contrast to the relative ease with which non-experts in cloud computing can build complex distributed applications that function reasonably well. We trace this difficulty in application building to the paucity of good systems research in robotics, and lay out a path toward enabling application building by centering usability in systems research in two different ways: privileging the usability of the abstractions defined in systems research, and ensuring that the research itself is usable by application developers in the context of evaluating it for its applicability to their target domain by following principles of realism, empiricism, and exhaustive explication. In addition, we make some suggestions for community-level changes, incentives, and initiatives to create a better environment for systems work in robotics.
\end{abstract}

\section{Introduction}

\subsection{The deployment gap in robotics}

Robotics is a fast-growing, multidisciplinary field with applications that are quickly leaping off the pages of science fiction into present day reality. The rapid spread of cheap, powerful mobile phones and their attendant sensor hardware, as well as the meteoric rise in the efficacy of machine learning techniques for perception, planning, control, and related problems, have left us on the cusp of an unprecedented golden age in robotics work, both in academia and industry.

We remain, however, \emph{just on the cusp} of that golden age. Self-driving cars are being tested all over the US but remain controversial and continually delayed, major equipment manufacturers and startups alike make loud noises about agricultural robots coming any day now, and people expectantly wait for their next package to be delivered by drone. Meanwhile, nothing much has really changed out in the field. The successful robots of the past three decades --- the industrial arms, the KIVA/Amazon Robotics warehouse robots, the iRobot vacuum cleaners --- continue to chug along, but successful new deployments, in particular those that create inarguable value, have been elusive \cite{bloomberg_2018}.

To some degree, this slow growth is to be expected. Robotics is hard. The physical world is riddled with inherent complexities, and many disparate strands of knowledge must be woven together to form a robotic system that does anything interesting, to say nothing of the exponential complexity of settings involving multiple robots coordinating with each other.

Despite the difficulty, the last decade has seen an unprecedented boom in both the founding and funding \cite{waters_2016, dowd_2019} of companies intending to bring robotics into wider use across a dizzying array of industries. Heartbreakingly, the vast majority of these companies have failed, including a number of high-profile examples run by luminaries from the robotics research community \cite{vanderborght_2019}. In particular, most of these companies appear to have failed not due to an inability to build the technology, but due to a failure to achieve product-market fit \cite{crowe_2018, schmelzer_2018} --- they were selling something that people didn't want. 

The founders of these companies were, by and large, roboticists rather than experts in the domains their companies were targeting. Their mode of failure suggests that the people who are best equipped to drive innovative new use cases for robotic technology are domain experts (in application domains like agriculture, healthcare, or construction, for example), since they are most able to anticipate the needs of their particular market.

\subsection{Application building}

In order to allow experts in application domains to drive the real-world adoption of robotics, we must enable \emph{fast application building}. An application here is a working end-to-end robot system artifact that can perform a task reliably and repeatedly in a real-world domain. The application does not have to be optimal or even very performant: it merely needs to work well enough to function as practical evidence for the utility of a robotic solution in the target domain.

Currently, application developers in robotics need to be deeply steeped in some mishmash of kinematics, dynamics, path planning, distributed systems, queuing theory, electrical engineering, and more. While one can certainly attempt to distribute the burden of knowledge across a team, at least one individual must have a working background in enough of the pieces to keep the whole thing together. Indeed, without such an individual it is nigh-impossible to figure out what kinds of problems are even tractable with current robotic technology. 

There are very few of these full-stack roboticists, and their scarcity limits the total number of attempts at building real-world robotic applications per unit time to a very small number. More attempts in a diverse set of application domains using distinct approaches, especially led by domain experts, would likely lead to more successes in real-world deployment. As such, it is incumbent on us as a field to prioritize enabling application building by technically sophisticated engineers who are not full-stack roboticists. This is where systems research comes in.

\subsection{The role of systems research}

The purpose of systems research, broadly defined, is to wrap complex machinery in human-usable abstractions that enable non-experts to build performant applications without having to understand every detail of the implementations underneath the interfaces they build on top of.

This is accomplished in a two-step process: first, the design and specification of usable abstractions that provide easy to reason about interfaces and guarantees for specific tasks, and second, the iterative refinement of implementations underneath these abstractions that provide better and better performance on whatever metrics the user cares about without breaking the abstraction's contract (though in practice these steps can be ordered in the opposite way, and they usually bleed into each other). Good systems work bridges the gap between abstract insight and real-world use cases by presenting just-right ``Goldilocks'' abstractions that are simultaneously simple enough to understand and use, powerful enough that real world applications can be built on top of them, and only loosely (if at all) tied to the vagaries of a specific implementation such that different implementations can be swapped in and swapped out under the abstraction layer.

An exemplar of a field where this is done right is distributed systems, which forms the academic foundation for cloud computing. Despite the dizzying array of hardware, software, and even fundamental physics concepts involved, anyone with basic computer science background can quickly learn to build and deploy a fairly complex distributed web application that scales to hundreds of thousands of users out of the box. It's as simple as writing an HTTP application on top of your favourite backend framework (Flask, Express, Revel) in your favourite language (Python, NodeJS, Go), interfacing it with a newly spun-up instance of the appropriate type of (relational, key-value) datastore (PostgreSQL, Redis), placing it behind a server (NGINX, Apache), and then just letting it run on a virtual machine. If you want to scale, you can replicate and/or shard your database, stick it behind a cache, automate the spinning up of more VMs for the application, stick that behind a load balancer, and so on. Every choice mentioned here turns on a small set of important tradeoffs, for example consistency vs. availability in the presence of network partitions and the richness of the query model vs latency for datastores, and can be made based on desired properties and workload assumptions for the system. What if you get it wrong? You need only to measure where you're deviating from your assumptions, revisit your tradeoffs, make some different choices, and redeploy.

This process is not trivial, though the proliferation of battle-tested hosted versions of all of the pieces involved by Google, Amazon, et al.\ has taken a lot of the pain out of it, but it isn't rocket science. Is it going to produce the \textbf{optimal} solution? No. Is it going to produce something \textbf{usable}? Very likely yes. Does it enable the creation of real-world applications that, but for the existence of systems that can compose in this usable way, would never have existed? Unquestionably yes. Done and working is better than vacuously perfect.

Aside from some reasonably healthy pieces of the ROS ecosystem (primarily authored and maintained outside of academia at places like Willow Garage, Clearpath, and OSRF) this sort of application building is very, very difficult, if not nearly impossible, in robotics today. Someone who wants to build their own serviceable (not even close to optimal or state-of-the-art) Amazon-style warehouse logistics application, for example, would likely not be able to do so without expert advice up and down the stack, despite the individual pieces of technology to do so being broadly within reach.  

\subsection{A way forward: Usability Squared}

We believe that in order to enable application building, we must center usability in robotics research in two different ways:

\subsubsection{Usable abstractions}
Systems research in robotics must prioritize designing abstractions that are intuitive for non-expert application developers to reason about and straightforward to use in building applications.

\subsubsection{Usable research papers}
While providing usable abstractions is essential, the research paper itself must also be usable in the sense of being accurately evaluable for its utility in a given target domain by a non-expert application developer. This involves using research methodology that privileges realism, empiricism, and exhaustive explication to demonstrate that the design choices made in the work and the tradeoffs exposed by the abstractions specified are the right ones for the domain or domains the paper is aimed at. 

In other words, we believe that the job of good systems research is to design and specify usable abstractions that are both \emph{intuitive} and \emph{powerful}, and the job of a good systems research paper is to validate the design choices made in specifying the abstraction and building its underlying implementation, thus making the research itself usable to an application developer. Both kinds of usability are essential in good systems work --- without either, the utility of the work in the real world is compromised. 

In the next two sections, we justify and elaborate on these two forms of usability.

\section{Usable abstractions}

If forced to choose, privilege the usability of the abstraction, in particular the intuitiveness of the interface to application developers and compositionality with other systems, over squeezing out the last drops of performance from the system or proving optimality. While it is often possible to squeeze greater performance out of less simple and intuitive abstractions, the increased complexity and resultant cognitive load generated often massively reduces the ability of application developers to easily make use of and compose them, thus preventing building applications that work correctly or, indeed, exist at all.

For an example of this phenomenon in action, we turn to distributed datastores, an area which recently witnessed the rise \cite{bailis2012probabilistically, helland2009quicksand} and decline \cite{muthukkaruppan_2010, lloyd2014don} of eventual consistency. Eventual consistency \cite{vogels2008eventually} promises better performance in the form of lower query latency and higher availability for distributed datastores via the mechanism of reducing the coordination required for each query. The tradeoff here is that the global state of the datastore is not guaranteed to be consistent --- this roughly means that if you write something to it, it will eventually be visible globally, but that is only guaranteed to happen as time goes to infinity. In the meantime, you may see inconsistent state in the system, with different reads returning conflicting values. This model is a significant departure from the strongly-consistent ACID (Atomicity, Consistency, Isolation, and Durability) semantics \cite{haerder1983principles} of classical databases.

ACID semantics and strong consistency roughly align with what human programmers intuitively expect from a datastore, and they simplify writing correct applications on top of systems that guarantee them \cite{drevets_2016}. Eventual consistency sacrifices that abstraction simplicity for performance. While eventual consistency datastores became quite popular in the late 2000s and early 2010s, they fell from grace because the tradeoff eventually came to be seen as not worth it \cite{yokota_2017}. Here's a representative quote from Google's paper about F1 \cite{shute2013f1}, their strongly-consistent high performance datastore for ads:

\begin{quote}
    ``We [also] have a lot of experience with eventual consistency systems at Google. In all such systems, we find developers spend a significant fraction of their time building extremely complex and error-prone mechanisms to cope with eventual consistency and handle data that may be out of date. We think this is an unacceptable burden to place on developers and that consistency problems should be solved at the database level.'' 
\end{quote}

Robotics is still a young field. When there are many more real-world deployments and application developers experienced in the basics of building robotic applications, we can start profitably experimenting with increasing the complexity of our abstractions, but until people are able to reliably use the simple stuff, this will be actively harmful. Indeed, without first optimizing for usability, we may never have enough data from application domains to even know what the quantitative metrics and use cases we should optimize for even are.

\section{Usable systems research papers}

An abstraction specified in new research work, no matter how usable in design and performant in implementation, can only be used by an application developer if they can confirm its applicability to their target domain by validating that the assumptions and design choices made correspond with the ground realities of a real-world deployment in their target domain, and that the tradeoffs exposed are the appropriate ones for said domain. 

In the spirit of Butler Lampson's classic \textit{Hints for Computer System Design} \cite{Lampson:1983:HCS:800217.806614}, we propose a few methodological principles in service of ensuring this second kind of usability in systems research.

\subsection{Principle 1: Target at least one specific real-world domain}

\begin{figure}
  \centering
  \includegraphics[width=130pt]{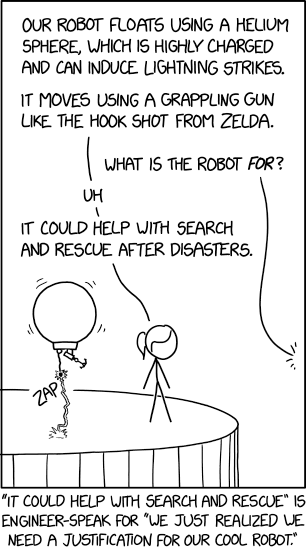}
  \caption{xkcd: New Robot by Randall Munroe (license: CC BY-NC 2.5) \cite{munroe_2012}}
  \label{fig:new_robot_xkcd}
\end{figure}

Good systems work comes from real-world problems. There is a great deal of work in robotics that seems to tack on an application domain as an afterthought, as the classic XKCD comic in Figure \ref{fig:new_robot_xkcd} \cite{munroe_2019} illustrates.

Having at least one real-world application domain ensures that at least application developers targeting that domain can use the work. In addition, having a concrete domain to compare against allows application developers targeting other domains to more easily evaluate the utility of the work for their domain.

\subsection{Principle 2: Make realistic assumptions and avoid unnecessary, unrealistic, and fanciful assumptions}

Good systems work is informed by real-world constraints. The assumptions that underlie systems research must align with the circumstances of (realized or hypothetical) real deployments of the application domain or domains the research is targeted at. 

\subsubsection{Necessary, realistic assumptions}

In order for research work to be applicable to a real-world domain, it must make assumptions that are fundamental to the operation of that domain, without which the work would be unrealistic and inapplicable. 

An important example of a necessary realistic assumption is that real robotic applications are \emph{always on and never stop}. A substantial portion of the power of robotic autonomy comes from its ability to facilitate the smooth, uninterrupted running of processes 24x7, and many existing and potential robotic applications don't (or won't) have a neatly-defined end state --- they would ideally just keep going, moving packages, building cars, and tending to fields until the end of time. Stopping, even for a few seconds, can be disastrously expensive. As such, it would behoove systems research that targets always-on domains to optimize for this assumption when possible.

Consider path planning. Classical path planning algorithms like A* do single-query one-shot planning of the whole path. In practice, real robots in always-on domains perform an iterative process of planning and replanning toward as they are given new goals within a somewhat but not maximally dynamic environment. Performing planning from scratch at every update cycle discards potentially useful state from the computation of prior plans. In the mid-to-late 2000s, there was a burst of research work on iterative multi-query path planning and ``anytime'' planning \cite{van2006anytime, souissi2013path} that sought to exploit this potentially useful state for faster and more optimal planning. It would make sense for systems that do path planning in quasi-dynamic environments to use these algorithms to, say, harness redundancies between the iterations to reduce average-case plan-update latency, which is the more important metric than worst-case cold-start plan-creation latency for always-on domains. For whatever reason, systems research in robotics tends to ignore this work, instead sticking with one-shot path planning techniques.

Perhaps in part due to this lack of uptake in systems research and real deployments, there is disproportionately little new work in this area relative to offline planning.

\subsubsection{Fanciful assumptions}

Fanciful assumptions are assumptions about the target domain, usually taking the form of very specific constraints, that are not supported by the ground realities of that domain. 

In multirobot systems work, there are a huge number of papers that focus on coordination given some specific, often unique unreliable communication model \cite{xiao2008asynchronous, fan2013virtual, cesare2015multi}. In almost all domains we care about, it is either possible to get quite reliable communication, for example by combining services from two consumer mobile broadband providers to get $99.999\%$ connection availability \cite{baltrunas2014measuring}, or it is not possible to get communication at all, for example in RF-denied nuclear disaster zones or in the deep sea. Non-military use cases requiring the use of some kind of ad-hoc mesh networking are largely limited to the exploration of caves and space, which collectively comprise a relatively small proportion of the domains that exist today. There is still a lot of work to do be done in domains where communication is reliable --- we have by no means solved multirobot coordination under those models --- but these much more realistic problems are often ignored. Assumptions like these should be strongly avoided.

\subsubsection{Beguiling assumptions that seem necessary but aren't}

There is a class of beguiling assumptions that are simple, intuitive, and seemingly useful, but in practice, at best, unnecessary and, at worst, actively harmful. Consider the assumption of deadlock-freedom in multi-agent path planning for warehouse domains. While it may seem entirely reasonable to want to guarantee that agents never deadlock, this guarantee is almost impossible without using totally centralized global planning, which severely limits scalability, and, crucially, is almost never a problem in practice --- companies using systems with no deadlock freedom guarantee see deadlock on the order of a few times a year even in very large deployments \cite{sankaran_knepper_2019}, and at that rate of occurrence it is better to simply have humans reset one of the robots after a timeout.

\subsection{Principle 3: Avoid irrelevant proofs and guarantees that are useless in practice}

Roboticists have a distinct affinity for theoretical proofs, even within systems work. While proofs of useful properties can certainly be beneficial in providing guarantees that make systems abstractions more usable, this fixation on proofs can be harmful in two ways: 

\begin{enumerate}
  \item If it slows down or prevents the publication of a practical contribution that can be empirically validated
  \item If a provable guarantee that is actually irrelevant clouds understanding of what metrics really matter and thus prevents the exploration of potentially profitable research directions
\end{enumerate}

A good example of the second phenomenon can be found in the literature around probabilistic, sampling-based planners such as the Probabilistic Roadmap (PRM)~\cite{kavraki1996probabilistic} and the Rapidly-exploring Random Tree (RRT)~\cite{Lavalle98rapidly-exploringrandom} planning algorithms. These motion planners rely on proofs of eventual probabilistic completeness that guarantee that some solution will be found as time goes to infinity. In practice, no robot has infinite time to wait, so it is common tradecraft to run RRT, for example, with a series of timeout-based restarts with the hope that different samples will produce a plan quicker. These restarts are seldom included in evaluations of systems using RRT, as noted by \citet{wedge2008heavy} in their excellent analysis of plan time distributions and restarts, and if they're mentioned at all it's perfunctory and not particularly well-explained, such as in this quote from the Forage-RRT paper \cite{keselman2014forage}:

\begin{quote}
    ``Moreover, any RRT reaching 10,000 nodes was restarted to improve the average planning time of all planners (empirically when an RRT grows too large, it will have trouble connecting to the goal, so it is better to restart).''
\end{quote}

For an application developer attempting to evaluate planners and planning systems, this sort of opacity around a crucial aspect of real-world deployment confounds their ability to make reasonable choices for their domain.

In addition, this sort of obfuscation may well harm future research in this area. There might be a motion planner that, for example, does not guarantee probabilistic completeness, but for all the domains we care about produces results faster than RRT (when tested empirically). In the current research paradigm, this planner's real-world superiority to RRT might not ever come to light. 

In general, the existence of some pervasive tradecraft secret like planner restarts that goes mostly unmentioned or unevaluated in the literature is a good heuristic for detecting that some guarantee being provided is unrealistic or useless --- there is usually an opening for good systems work to be found in these situations.

\subsection{Principle 4: Explicitly justify design choices with reference to counterfactual designs}

Given that the point of a systems paper is to justify the design choices made in the research described, it is essential to explicitly consider counterfactual options, the roads not taken, to justify why the choices made were the correct ones. This not only helps evaluate work in comparison with related work in the area, and in general better validate the reasoning behind design choices, but also helps the application developer distinguish design choices that are essential to ensure the proper functioning of the system from design choices that can be safely modified depending on the specifics of a particular domain or implementation. This can be the difference between an application developer incorrectly seeing some research as incompatible with their domain and that same developer profitably using the core ideas of the work while modifying things on the periphery to achieve compatibility with their domain.

Computer architecture papers often do a very good job of this kind of design space exploration and justification. Here's a quote from the abstract of the Q100 paper, which proposes an architecture for a specialized Database Processing Unit (DPU) (\citet{wu2014q100}):

\begin{quote}
  ``This work explores a Q100 design space of 150 configurations, selecting three for further analysis: a small, power-conscious implementation, a high performance implementation, and a balanced design that maximizes performance per Watt. We then demonstrate that the power-conscious Q100 handles the TPC-H queries with three orders of magnitude less energy than a state of the art software DBMS, while the performance-oriented design outperforms the same DBMS by 70X.''
\end{quote}

\begin{figure*}
  \centering
  \includegraphics[width=\textwidth]{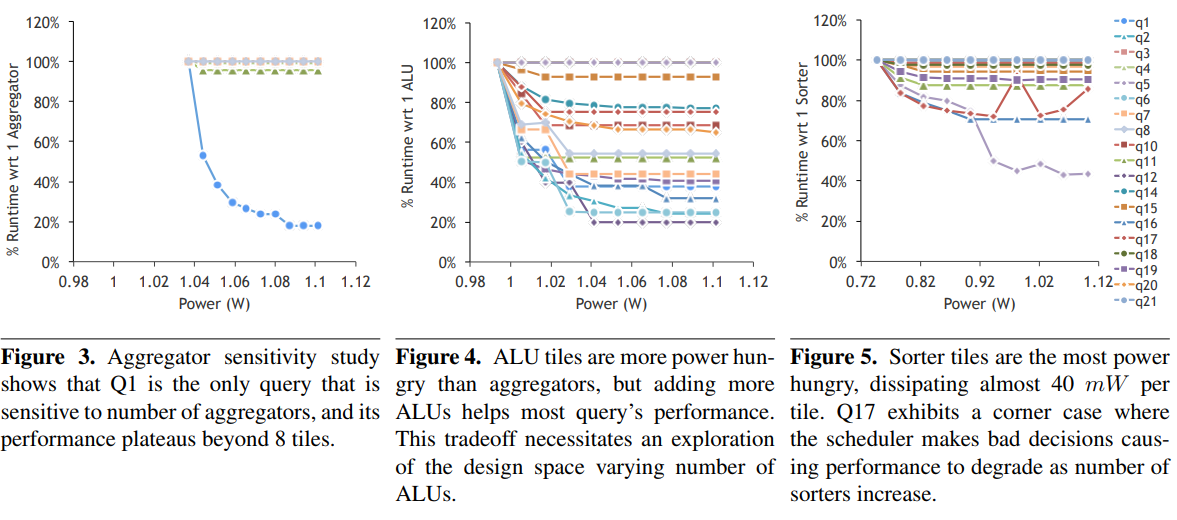}
  \vspace{-5mm}
  \caption{Design space exploration graphs from \citet{wu2014q100}}
  \label{fig:q100long}
\end{figure*}

The graphs in Figure \ref{fig:q100long}, which are Figures 3, 4, and 5 from \citet{wu2014q100}, detail some of the analyses they ran as part of their design space exploration in which they vary the number and connectivity of various component types.

\begin{figure}
  \centering
  \includegraphics[trim={0 0 0 0.4cm},clip, width=\linewidth]{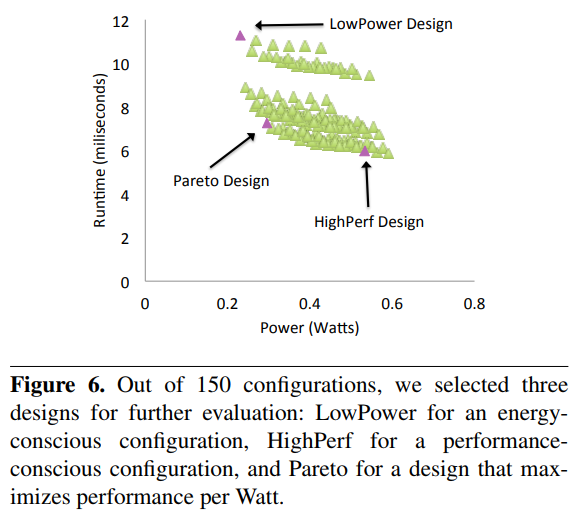}
  \vspace{1mm}
  \caption{Performance relative to power consumption of 150 different configurations from \citet{wu2014q100}}
  \label{fig:q100perfpower}
\end{figure}

The graph in Figure \ref{fig:q100perfpower}, which is Figure 6 from \citet{wu2014q100}, charts the performance relative to power consumption of their 150 different configurations, highlighting the three they chose for further study.

\subsection{Principle 5: Make your tradeoffs explicit and empirically explore the tradeoff space}

The design choices made by systems will fix some set of parameters and expose other sets of parameters as tunable tradeoffs. These tradeoffs, which act as knobs that application developers can twiddle, need to be explicitly highlighted, motivated, and empirically explored in order to allow for evaluation with respect to real-world conditions in a target domain.

An example of a tradeoff exposed in robotics is discretization granularity. Discretizing space is a common strategy for path planning, especially in the multirobot domain~\citep{honig2019persistent, alami1998multi, wagner2015subdimensional}. Discretization represents a set of tradeoffs against plan-optimality in the real (continuous) world, including speed of computation, simplicity of algorithm, and ease of implementing occupancy-based safety guarantees. These tradeoffs are rarely explicated in direct ways, and the tradeoff space is almost never empirically explored in order to justify the design decisions made and parameters selected. How does finer and finer discretization affect compute time in various realistic settings? How close to optimal do you get with reasonably granular discretization? Is there some optimal point on the tradeoff graph where the computation is quick enough for use on real robots and the solutions generated are close to optimal, with diminishing returns for finer granularity? These questions are incredibly relevant for both real-world use and future research directions, but are almost never answered. 

An example of this in action in database systems is the fundamental tradeoff between strength of consistency guarantees and query latency, as explicated by Dan Abadi in his PACELC principle \cite{abadi2012consistency}, which is an extension of Eric Brewer's CAP theorem \cite{Brewer:2000:TRD:343477.343502, brewer2010certain} that famously limits a distributed datastore to two out of the three of strong Consistency, constant Availability, and Partition tolerance. Database systems such as Cassandra allow you to choose different levels of consistency \cite{cassandraconsistencydoc}, with higher levels resulting in higher latency, and these systems have been subject to comprehensive of studies of performance along the tradeoff axis \cite{haughian2016benchmarking}.

\subsection{Principle 6: Test till you break, scale till you fail}

\begin{figure}
  \centering
  \includegraphics[width=\linewidth]{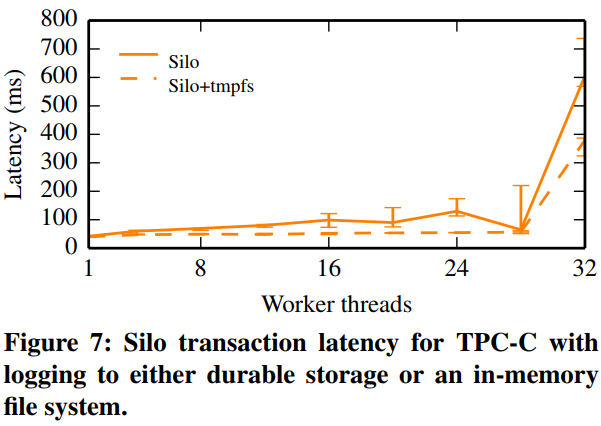}
  \vspace{1mm}
  \caption{Transaction latency graph from \citet{tu2013silo}}
  \label{fig:silo}
\end{figure}

Don't just publish a graph demonstrating linear performance scaling for, say, a multirobot path planning system from 1 to 20 robots --- graph your results past the point where you stop being able to scale. This is incredibly important for pointing to where future work needs to improve, and it helps people using your system bracket the range of reasonable performance for their needs. Database papers often do this particularly well, with Figure \ref{fig:silo}, which is Figure 7 from the SOSP 2013 Silo paper \cite{tu2013silo}, serving as a good example. 

The graph demonstrates that latency performance is roughly unchanged as you scale the number of worker threads until a massive spike past 28 threads. This result allows an application developer working in, for example, a target domain with a requirement for latency lower than 200ms to decide whether Silo is appropriate for them based on whether they need greater or fewer than 28 threads to satisfy their throughput requirements. 

\subsection{Principle 7: Use or make open, public benchmarks and baselines, ideally based on real-world workloads.}

The machine learning, computer systems, and databases communities make rapid progress due to the existence of open datasets (ImageNet \cite{deng2009imagenet}), load generation tools (YCSB \cite{cooper2010benchmarking}), and benchmarks (TPC-C \cite{Leutenegger:1993:MST:170035.170042}). In addition, in the latter two communities, it is common to try and model real-world workloads or replicate them exactly using traces sourced from industry \cite{pucher_2015, Cortez:2017:RCU:3132747.3132772}. 

Systems work in robotics often relies on either \textit{proof by video}, where a cherry-picked short sample of footage of the system in action is used as evidence of efficacy, at worst, or homespun ill-defined metrics at best. This is insufficient. Without common benchmarks and baselines, it is impossible to evaluate work relative to other work in a principled manner, and for systems work, this means that it is effectively impossible to divine the correctness of the design choices made in any specific paper. Aside from overlap with the machine learning and computer vision communities (from which datasets like KITTI \cite{geiger2013vision} have emerged), robotics largely does not produce or utilize such benchmarks \cite{del2006benchmarks}, with the exception of somewhat infrequent and inconsistently targeted competitions \cite{Anderson:2011:RRC:1972812.1972815}. This is true even for problems where there should be strong incentives to make apples-to-apples comparisons, including path planning (single and multirobot) and task allocation for multirobot systems. There is some evidence that this state of affairs is changing, with the help of work like Nathan Sturtevant's planning benchmarks \cite{sturtevant2012benchmarks}, but it is not changing quickly enough.

In particular, the use of end-to-end benchmarks based on simulated workloads that model real-world deployments, for example the Asprilo warehouse logistics problem generator \cite{asprilo_2018} developed by Gebser et al \cite{gebser2018experimenting}, would greatly improve the ability of application developers to judge the relevance of research to their target domain.

\subsection{Principle 8: Exhaustive explication --- no tricks up your sleeve --- to enable replication}

As noted above, robotics research is plagued by the problem of proofs by video. It is often impossible to replicate these videos in academic settings, where some system flakiness is tolerable, much less in the high-reliability low-error-tolerance world of industry. This lack of replicability stems from the many undocumented patches, hacks, and simplifying assumptions used to get a robotic system to run that are passed down only via intra-lab oral tradition, if at all, and it fatally destroys the value of the research by preventing both further academic exploration and validation by independent groups as well as any kind of real-world deployment.
It is thus incumbent on systems researchers to ensure that any work they produce is reproducible. There has been much recent discussion of the reproducibility crisis in robotics and methodological principles for conducting reproducible research in our field \cite{towardsreprobots, rramrep}, and the first (though by no means the last) step towards reproducibility is to document every detail of how the system was made to work, from physical specifications to operating system versions to lighting conditions. While conference papers have space constraints, it is easy to put a supplementary document on arXiv or a similar archival service.

This principle very much does not supplant or subsume all the work being done on reproducibility in the sciences and engineering in general, and in robotics in particular. We recommend anyone working on robotic systems keep abreast of the latest best practices in this area.

\section{Community-level suggestions for promoting good systems work}

While we certainly hope that individual research papers following the principles outlined will lead to better work overall, the overall paucity of good systems research in robotics can only be rectified by consistent community-level acceptance and encouragement of this work. While the major robotics conferences pay lip service to wanting more systems work, explicitly welcoming and endowing awards for submissions of this type, word on the street is that the work is often not seen as core research, and the standards for inclusion are often inconsistent. People have very different and often conflicting opinions of what a systems research project or a systems paper even is. If we want to enable application building, we must make space within the community for the systems work that forms the foundation for it.

In the hope of beginning that process, we present a set of community-level suggestions for creating a better environment for good systems work.

\subsection{Suggestion 1: Recognize that finding a useful new way of structuring a problem is a first-class research contribution}

The bread and butter of systems work is reorganizing a problem around some key insight or set of insights such that it becomes more tractable. Once this is accomplished, the actual solution may seem easy to conceptualize and implement. This should not serve to diminish the validity of the research contribution of the work --- indeed, one hallmark of great systems work is that it repurposes preexisting abstractions and components to solve new problems, reducing redundant work and exposing fundamental shared structure. It is easy to dismiss the value of restructuring a problem domain, and this impulse must be fought vigorously.

\subsection{Suggestion 2: Value ``incremental'' systems work}

Work that builds on pre-existing research in ways that are not revolutionary but still substantive, improving performance on the same benchmarks and baselines or providing some new useful feature, is necessary and important, both for ensuring that exciting work from academia reaches the threshold of real-world acceptable performance and for ensuring that new work that purports to be a revolutionary is not worse than previous generation work with incremental modifications. If the latter situation is not detected because no one ever bothered to build the optimizations, then researchers might go down suboptimal research paths due to the incorrect belief that the prior pathway could never reach the performance that the new one does.

In computer systems and databases, there is a strong tradition of work that preserves the same programmer-facing abstraction (or very close to it) while changing the implementation in intelligent ways to significantly improve performance. The methods behind these sorts of improvements must be recognized as first-class research contributions to ensure that work like this can exist in robotics. 

\subsection{Suggestion 3: Incentivize the creation of open, public benchmarks and workload datasets}

As noted in the principles, it is essential that researchers test their systems on public benchmarks. As a community, we should incentivize the creation of such benchmarks and the publication of workload datasets with guaranteed publication slots, special awards, and cold, hard cash via the creation of some kind of benchmark fund or prize. These carrots should be complemented by the stick of subjecting to strict scrutiny and possible publication denial papers that either don't use appropriate common benchmarks, don't release their own benchmarks, or both.

\subsection{Suggestion 4: Incentivize frontier-illuminating papers --- systems papers that try to build some arbitrary application in a principled way using state-of-the-art research}

A combination of the fractal nature of robotics as a field and the preponderance of work that doesn't follow the principles outlined above makes it hard to see the frontier of our ability to build systems that solve any specific real-world problem. Competitions \cite{Anderson:2011:RRC:1972812.1972815} like the Amazon Picking Challenge \cite{correll2016analysis} have been the primary method of illuminating these frontiers, but have the downside of being infrequent and industry-controlled. As a result, we believe that we should build a program of practical frontier papers where groups are funded to solve a rotating set of real-world problems --- for example clearing a cluttered home or the monitoring and harvesting of a specific crop in a greenhouse --- using existing research, then report on their methods and results on a set of predefined open benchmarks. This sort of work will also serve to illuminate the practical frontiers of the constituent subproblems of these real-world problems, tying performance to real-world workloads. 

The Integrated Intelligence for Human-Robot Teams paper by \citet{oh2016integrated} is a good model for this kind of work.

\section{Conclusion}

We believe that making space for more and better systems research that follows the principles we've outlined will not just help us make progress toward the goal of seeing robots widely deployed in the real world, but also benefit academia immensely by providing new problems motivated by experiences that application developers have in their particular domains, opening up a large pool of potential new sources for funding, and motivating a larger and more diverse set of people to work in robotics after seeing it in their daily lives. This has been the experience of research communities in other computer science and engineering fields, most prominently of late in machine learning.

There are millions of potential application developers out there waiting for us to unlock their ability to reimagine the world we live in. We are in the pre-Apple II days of robotics, tooling away with our expensive research toys with only baroque industrial deployments to prove the worth of our field. It is our duty to bring the power of robotics to the people, and we can scarcely imagine the depth of the ingenuity that doing so will reveal.

\section*{Acknowledgments}

Thanks to Wil Thomason and Dylan A. Shell for their participation in initial discussions leading up to this paper, as well as to Natacha Crooks, Adrian Sampson, Chris De Sa, and A. Feder Cooper for suggesting systems papers worth referencing. 

In addition, we'd like to thank Christopher Leet for his input on the structure of the paper, as well as Wil Thomason, Tom Magrino, Sowmya Dharanipragada, Danny Adams, Alexa VanHattum, Kate Donahue, Gregory Yauney, and A. Feder Cooper for reading drafts of it.

This paper is based on work partly supported by the National Science Foundation under Grant No. 1646417.  We are grateful for this support.

%% Use plainnat to work nicely with natbib. 

\bibliographystyle{plainnat}
\bibliography{references}

\end{document}